\documentclass[switch]{article}
\usepackage[preprint]{corl_2024}
\usepackage{graphicx}
\usepackage{dblfloatfix}
\usepackage{url}
\usepackage{wrapfig}
\usepackage{booktabs}
\usepackage{multirow}
\usepackage{lineno}
\usepackage{placeins} 
\usepackage{afterpage} 
\usepackage{natbib}
\usepackage{pgf}
\providecommand{\mathdefault}[1]{#1}

\newif\iffigs
\usepackage{capt-of}  
\usepackage{cuted}
\usepackage[font=small,labelfont=bf]{caption}
\usepackage{subcaption}

\usepackage[fleqn]{amsmath}
\usepackage{float}
\usepackage{listings}
\usepackage{setspace}
\usepackage{mathrsfs}
\usepackage{amssymb}
\usepackage{nicefrac}
\usepackage{algorithm}
\usepackage{algorithmicx}
\usepackage[noend]{algpseudocode}

\usepackage{xcolor}
\usepackage{listings}
\usepackage{textcomp}
\definecolor{backcolour}{rgb}{0.95,0.95,0.92}
\lstdefinestyle{mystyle}{
    backgroundcolor=\color{backcolour},   
    basicstyle=\ttfamily\footnotesize,
    breakatwhitespace=true,         
    breaklines=true,                 
    captionpos=b,                    
    keepspaces=true, 
    keywords={},
    showstringspaces=false,
    showtabs=false,                  
    tabsize=2
}
\usepackage{pythonhighlight}
\usepackage{flushend}

\makeatletter
\newcommand\fs@spaceruled{\def\@fs@cfont{\bfseries}\let\@fs@capt\floatc@ruled
  \def\@fs@pre{\vspace{0.4\baselineskip}\hrule height.8pt depth0pt \kern2pt}%
  \def\@fs@post{\vspace{-0.4\baselineskip}\kern2pt\hrule\relax\vspace{-12pt}}%
  \def\@fs@mid{\kern2pt\hrule\kern2pt}%
  \let\@fs@iftopcapt\iftrue}
\makeatother

\usepackage{lipsum}
\usepackage{changepage}

\usepackage{etoolbox}

\title{\LARGE \bf Towards Forceful Robotic Foundation Models: a Literature Survey}
 \author{William Xie and Nikolaus Correll\thanks{$^{1}$All authors are with the University of Colorado at Boulder, Boulder, CO. Corresponding email: {\tt\footnotesize wixi6454@colorado.edu}}}
\begin{document}
\maketitle
\begin{abstract}
This article reviews contemporary methods for integrating force, including both proprioception and tactile sensing, in robot manipulation policy learning. We conduct a comparative analysis on various approaches for sensing force, data collection, behavior cloning, tactile representation learning, and low-level robot control. From our analysis, we articulate when and why forces are needed, and highlight opportunities to improve learning of contact-rich, generalist robot policies on the path toward highly capable touch-based robot foundation models.
We generally find that while there are few tasks such as pouring, peg-in-hole insertion, and handling delicate objects, the performance of imitation learning models is not at a level of dynamics where force truly matters. Also, force and touch are abstract quantities that can be inferred through a wide range of modalities and are often measured and controlled implicitly. We hope that juxtaposing the different approaches currently in use will help the reader to gain a systemic understanding and help inspire the next generation of robot foundation models.

\end{abstract}

\vspace{-5.0mm}

\section{Introduction} \label{sec:intro}

With the world population in the industrialized part of the world shrinking \citep{pop_collapse}, the need for generalist robotic systems capable of caring for an aging population and filling in gaps in the working population is rapidly growing. Emerging rapidly in response is a new industrial sector that focuses on humanoid robots---robots that can seamlessly integrate into existing workflows due to their human-like shape and sensor configuration. Concurrently, so-called ``robot foundation models'' \citep{rt1, openx, openvla, pi0, gemini_robotics} have demonstrated zero-shot autonomy for a series of dexterous manipulation tasks via combining imitation learning with the world knowledge provided by large language models \citep{brown2020gpt3}.

Force and touch are critical modalities for robotic systems that interact in the real world. Being able to control not only the robot's position, but also the force it exerts on its environment, is critical for manipulating delicate objects \citep{dg}, human-robot interaction, and contact-rich manipulation for assembly \citep{watson2020autonomous}. However, current robot foundation models focus exclusively on visual input and position control. Relying on position alone might not be sufficient to achieve true dexterity, as small errors in position may lead to large errors in force in stiff systems. And because force, the second derivative of position (via Newton's second law of motion $F=m\ddot{x}$), represents a richer, higher-frequency representation of motion, focusing only on position may not be sample-efficient or capture high-frequency information during imitation learning. Yet, why and how forces should be employed during learning remains still unclear, in particular, as many of the benefits of force control can be gained using implicit techniques ranging from impedance control to mechanism design. 

In this paper, we review recent efforts to extend end-to-end robot learning to force and touch sensing while situating this work in the larger context of force control and tactile sensing in robotics. Here, we specifically focus on transformer \cite{vaswani2017attention} and diffusion-based \cite{dp} end-to-end learning methods, which, due to their favorable scaling properties, have the potential to integrate with generalist foundation models. We hypothesize that the next generation of robot foundation models will require force and torque input and output and hope that the synthesis of the existing literature in this survey will contribute to their design.  

We begin this survey with a background on the human sensing apparatus, force control in robotics, and the field of policy learning (Section \ref{sec:background}). After providing an overview of the reviewed works by force and timescale (Section \ref{sec:review}), we discuss work in forceful end-to-end learning with regard to data collection (Section \ref{sec:data_collection}), ways of generating force trajectories (Section \ref{sec:action_space}), and utilizing and scaling representation of force data (Section \ref{sec:policy_learning}). We conclude with a discussion on the advantages of forceful policies and future directions. 

\section{Background}\label{sec:background}
In this section we provide context for force and touch sensing, the latter of which we use synonymously with tactile sensing, and robot policy learning. We also situate tactile robot policies in the broader field of robot learning and the rise of large scale datasets and robot foundation models.

\subsection{Human Tactile Sensing and Proprioception Apparatus}
We precede this survey with a brief overview of the human tactile and force sensing apparatus. This is important for two reasons: (1) the differentiation of the human sensing system suggests that tactile-based dexterity relies on a data with a wide variety of spatial resolution, bandwidth, and signal dynamics; (2) the impact of specific sensory information on makespan and precision for a variety of manipulation is well understood in humans, thereby possibly informing the design of robotic systems. 

The human tactile sensing system relies on specialized \emph{mechanoreceptors} located within the skin, each adapted to detect specific types of mechanical stimuli. These mechanoreceptors are categorized into four primary types: Fast-Adapting Type I and II (FA-I and FA-II), and Slowly Adapting Type I and II (SA-I and SA-II). Each type exhibits distinct physiological and functional properties that contribute to the sense of touch. FA-I mechanoreceptors, associated with Meissner corpuscles, are predominantly found in the glabrous (hairless) skin of the fingertips. They respond to low-frequency vibrations in order of 20--200Hz \citep{wolfe1988sensory} and dynamic skin deformations, playing a crucial role in detecting textures and slip during object manipulation. Their receptive fields are small and well-defined, allowing for precise spatial resolution \citep{johansson1983tactile,johnson2001roles}. FA-II mechanoreceptors, linked to Pacinian corpuscles, are distributed throughout the hand. These receptors are particularly sensitive to high-frequency vibrations up to 1500Hz \citep{wolfe1988sensory} and sudden changes in pressure. They contribute to the perception of fine textures and tool use. Unlike FA-I mechanoreceptors, FA-II receptors have large and diffuse receptive fields, enabling them to detect distant vibrations \citep{johnson2001roles, mountcastle2005sensory}. SA-I mechanoreceptors, associated with Merkel cell-neurite complexes, are concentrated in the fingertips and specialize in detecting sustained pressure and fine spatial details. They are essential for form and texture discrimination due to their small and well-defined receptive fields \citep{johansson1983tactile}. SA-II mechanoreceptors, connected to Ruffini endings, are evenly distributed across the glabrous skin of the hand. They are particularly sensitive to skin stretch and sustained pressure, contributing to the perception of hand shape and finger position. Their large and diffuse receptive fields aid in proprioceptive feedback \citep{vallbo1984properties}. 

\emph{Proprioception} refers to the body’s ability to sense its position, movement, and the forces exerted by its limbs and joints without relying on external sensory input (e.g., vision). This sensory feedback is crucial for maintaining posture and executing coordinated, precise movements, such as in dexterous manipulation. Key proprioceptive sensors involved in detecting joint torque and muscle activity are muscle spindles, Golgi tendon organs (GTOs), and joint receptors. Muscle spindles \citep{banks2021secondary} are embedded within muscles and detect changes in muscle length and the rate of stretch. When a muscle is stretched, muscle spindles send signals to the brain to inform it of the muscle's length and how fast it is changing. This helps the brain adjust muscle activity to prevent overstretching and maintain proper muscle force during movements. Golgi Tendon Organs (GTOs) \citep{jami1992golgi} are found in the tendons. GTOs sense the tension or force exerted by muscles. When the tension exceeds a certain threshold, GTOs inhibit further contraction to prevent muscle damage. This feedback is essential for regulating joint torque during activities that require force precision. Joint Receptors \citep{tuthill2018proprioception} are located in the joint capsules and ligaments and detect changes in joint position, movement, and stretch. They provide feedback to the brain about the angle and motion of joints, which is essential for maintaining stable postures and controlled movements.

Researchers have also studied the impact of the absence of certain sensing modalities.  For example, local anesthesia of tactile mechanoreceptors \cite{johansson1984roles} show severe impairement of dexterity, but also demonstrates that the task can eventually be completed using visual feedback and proprioception alone. This is also impressively demonstrated in a video showing a women striking a match with and without local anesthesia numbing the fingers \cite{youtube_video_0LfJ3M3Kn80}. Regarding proprioception, \cite{tuthill2018proprioception} reports of a case of a young man who was able to relearn muscle control after a neurological disease disabling his proprioception system.

In robotics, the term proprioception usually refers to joint encoders and torque sensors that are internal to the robot, whereas tactile sensors fall into the category of exteroception, i.e. external to the robot. We note that the transition between force/torque and tactile sensing is quite fluent as these quantities are mechanically linked and that while proprioception is well developed, robotic tactile sensing generally trails human capabilities in terms of information density and the ability to measure shear forces. Here, the main challenges are less the existence of appropriate measures, but integration and manufacturing \cite{mcevoy2015}.

\subsection{Force Control}\label{sec:bg_force_control}
In order to better understand the relationship between position and force and its implications for robot policy learning, we briefly review robotic force control \cite{siciliano1999robot}. A robot linkage with a Jacobian matrix $J \in \mathcal{R}^{6xn}$, the partial derivatives of all of its \textit{n} joints $q \in \mathcal{R}^n$ to the end-effector pose $x \in \mathcal{R}^6$, can control its end-effector velocity $\dot{x}$ via its joint velocity $\dot{q}$ using 
\begin{equation}
\dot{x}=J\dot{q}.\label{eq:fwdk}
\end{equation}
We can use the same Jacobian to compute the necessary joint torques $\tau \in \mathcal{R}^n$ as 
\begin{equation}
\tau = J^TF\label{eq:fwdd}
\end{equation}
where $F$ is a spatial wrench consisting of three translational forces and three torques around the principal axes \citep{correll2022introduction}. 

If a robot does not provide the ability to control joint torques directly, we can employ impedance control \citep{hogan1985impedance} to command a relationship between position and force:
\begin{equation}\label{eq:impcontrol}
    F = M \ddot{x} + D \dot{x} + K (x - x_d)
\end{equation}
Here, $\dot{x}$ and $\ddot{x}$, the current velocity and acceleration, are inputs to a \emph{virtual} spring-mass-damper system computes the force that results from the difference between current pose $x$ and desired pose $x_d$. In other words, if the robot end effector had mass $M$ and was attached to pose $x_d$ with a virtual spring and damper, moving it to $x$ would exert the wrench $F$ due to the spring coefficient $K$ (Hooke's law). Letting the end-effector go would let it snap back in place, oscillating like a real spring given its damping coefficient $D$. If we ignore mass and damping ($M=0, D=0$), we would control only stiffness, also known as \emph{compliance control}. 

Alternatively, we can solve (\ref{eq:impcontrol}) for $\ddot{x}$ and integrate the velocity and pose numerically to obtain the necessary displacement to exert force $F$. This is known as \emph{admittance control} \citep{keemink2018admittance}. As force readings might not be available along all degrees of freedom, thereby preventing closed-loop control around actual force, there exist also hybrid control schemes that control force only along some principal axes and use position control for the other dimensions.  

In the context of policy learning, approaches that learn from poses and implement position or velocity control require implementing a solution to inverting (\ref{eq:fwdk}) (also known as differential inverse kinematics), whereas controllers that aim at imitating end-effector wrenches will need to provide solutions to invert (\ref{eq:fwdd}) (inverse dynamics), possibly via the detour of admittance control. With this in mind, some policy learning frameworks also directly record joint-space positions, velocities and torques. While this sacrifices transferring policies between robots with different kinematics such as in the Open-X dataset \citep{openx} which explicitly records trajectories in relative end-effector coordinates, imitation in joint space is less prone to singularities and numerical problems that arise from inverse kinematics/dynamics and admittance control. 

Policies can also learn impedance control for a variety of contact-rich manipulation tasks in  \citep{buchli2011learning,abu2020variable,zhang2021learning,park2023lstm}, where they help position-based frameworks to generalize better. 

\subsection{Policy Learning}\label{sec:policy_learning_background}

Imitation learning (IL) is a subfield of machine learning where an agent learns to perform tasks by mimicking expert demonstrations. Unlike reinforcement learning (RL), which relies on trial-and-error with a reward signal, IL directly infers the desired behavior from demonstrations and trains a model to match the demonstrations in a least-squares manner. IL is particularly useful in robotics, autonomous driving, and interactive AI applications where defining a reward function is difficult or unsafe. The two primary types of IL are (1)  behavior cloning (BC), supervised learning applied to mapping states to expert actions, and (2) inverse reinforcement learning (IRL), learning a reward function that explains expert behavior and then optimizing it using RL \citep{ng2000algorithms}.

While reinforcement learning-based approaches suffer from the curse of dimensionality, which makes many real-world learning problems intractable, BC is effective with comparatively fewer and sparser demonstrations. However, BC is then very brittle during inference in situations that have not been part of the initial training set or are ``out-of-distribution''. Techniques like DAgger (Dataset Aggregation) \citep{ross2011reduction} and Guided Policy Search \citep{levine2013guided} attempt to bridge IL with RL, helping agents recover from errors and improve performance beyond expert capabilities.

Recent advancements in transformer and diffusion model architectures have reshaped IL, reducing the reliance on RL-based optimization. Models like Decision Transformer (DT) \citep{chen2021decision} and Trajectory Transformer \citep{janner2021offline} represent policy learning as a sequence modeling problem that generates actions much like a canonical text-based transformer generates characters. Instead of explicitly optimizing a reward function, these models treat demonstrations as a language-like sequence and generate future actions in an auto-regressive fashion. Unlike the classical formulation of a Markov Decision Problem, in which each state depends only on the previous state, self-attention in transformers \citep{vaswani2017attention} enables conditioning upon a large number of previous states, thereby allowing learning agents to discover implicit recovery mechanisms instead of blindly mimicking the expert.   

Diffusion models, originally developed for image generation (e.g., DALL-E 2 \citep{ramesh2022hierarchical}, Stable Diffusion \citep{rombach2022high}), have been adapted for IL by learning to denoise suboptimal trajectories into expert-like behaviors. Diffusion \citep{janner2022planning} applies noise to expert demonstrations and learns to refine them, resulting in smooth, human-like control policies. Here, the diffusion process ensures consistency across the entire trajectory. Diffusion policy \citep{dp} outperformed behavior cloning and RL baselines in robotic manipulation tasks by capturing uncertainty in demonstrations. Diffusion policy is a special case of flow-matching \cite{lipman2022flow}, which learns a velocity field that transforms a zero mean, variance one normal distribution into a target distribution. In this context, transformers are used to encode sensory data and text commands, which can then guide the diffusion / flow-matching process using feature-wise linear modulation (FiLM) encoding \cite{perez2018film}. 

At the same time, the transformer architecture has also revolutionized policy learning by using large language models for code generation. Based on Google's PalmE model, \citep{ahn2022can} has demonstrated a new level of open-world reasoning for mobile manipulation by choosing from LLM-generated suggestions for the next best learned policy \citep{jang2022bc} using learned value functions \citep{kalashnikov2021mt}. In our own work, we have chosen a similar approach to combine the open-world knowledge of an LLM to tune a variety of hand-coded controllers to manipulate delicate objects \citep{dg}. While seemingly at odds with action-generating transformers, these two approaches are starting to fuse in Vision Language Action models (VLA) that combine large pre-trained VLMs with transformer- \citep{kim2024openvla} or diffusion-based action heads \citep{wen2025tinyvla}, or even directly generate trajectory end-points from a VLM as in Gemini Robotics \citep{gemini_robotics}. VLMs can also be fine-tuned to explicitly reason about the properties of physical objects \citep{dg} or spatial concepts \citep{gemini_robotics} to increase their utility during manipulation tasks. 

In this article, we primarily review works which leverage data-driven methods for learning robot motion generation, of which diffusion \citep{dp} and transformer \citep{rt1, bet, act} based architectures are most common. This area of robot policy learning is often described as \textit{end-to-end robot learning}, in contrast with approaches which compose modules responsible for planning, vision, and control, as it is a method for learning a direct mapping from raw inputs to robot actions \citep{bekris_state_motion_generation_2024}, e.g. sensing-to-torques. However, end-to-end is a nebulous and often uninformative term, especially as 1) there is a wide performance gap between end-to-end methods which leverage large pretrained models and those which train on limited demonstrations \citep{lin2025datascalinglawsimitation} and 2) such methods do not ever capture the full robotics spectrum (nor should they ever be expected to, many will argue), requiring first and last mile help to go from long-horizon planning to precise force control.

Still, however imperfect, we use the term \textit{robot policies} to describe the resultant state-action mappings produced from end-to-end (synonymous with data-driven or implicit) learning methods \citep{bekris_state_motion_generation_2024}. 

\subsection{Touch and Force Sensing}\label{sec:touch}
In robotics, the sense of touch has been incorporated across a spectrum of embodiments, scale, and applications. Within manipulation alone, touch is deployed for grasping, in-hand manipulation and localization, object pose estimation, and object reconstruction \citep{Howe01011993_tactile_review, dahiya_tactile_review_2010, tactile_review_recentprogress,tactile_review_suomalainen_survey_2022,bhirangi_tactile_review_2024, tactile_review_li_comprehensive_2024}. 

\begin{figure}[!htb]
\includegraphics[width=\textwidth]{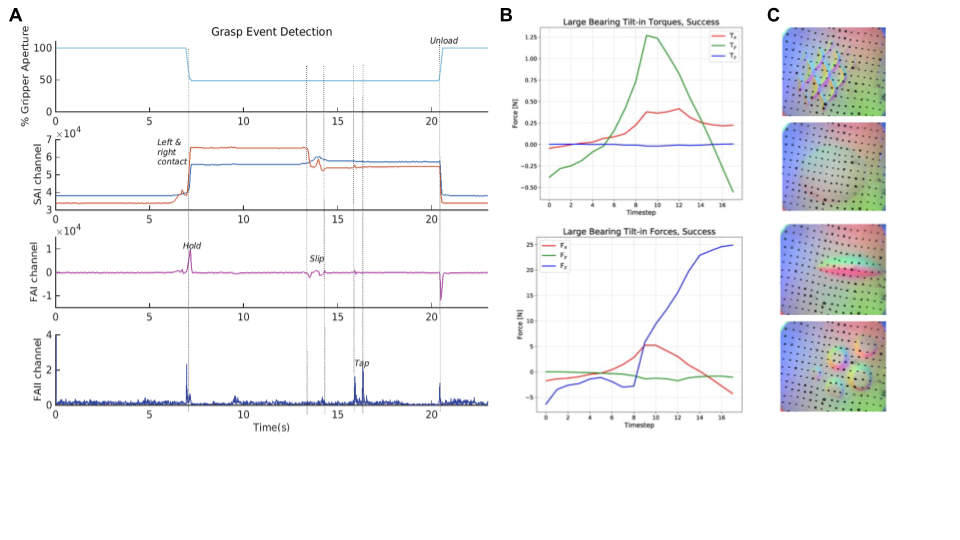}
\vspace{-55px}
\caption{A: Differential tactile data during a sequence of grasping events. Gripper aperture (top row), pressure sensing in the left and right finger tip (2nd row), the derivative of the pressure signal (3rd row), and accelerometer at the wrist (4th row), from \citep{patel2017improving}. B: Force (top) and torque (bottom) data over time during a successful bearing insertion from \citep{watson2020autonomous}. C: High-resolution tactile information from a GelSight sensor from \citep{calandra2017feeling}.\label{fig:touchdata}}
\end{figure}

The hardware space of tactile robot manipulation is quite heterogeneous, ranging from 1) normal and shear force sensing on end-effector finger-mounted force sensors, 2) force-torque sensing at the wrist joint preceding the end-effector, 3) extrapolation of end-effector external wrenches to joint motor torques, 4) finger-mounted optical sensors (synonymous with visuo-tactile finger sensors) that capture high-resolution contact deformation imaging, 5) finger-mounted magnetometer, piezoelectric, and capacitive sensors that capture similar information as 4) but with lower dimensionality and modality-specific accommodations, and 6) assorted novel sensing methods, such as robot skins \citep{hughes2015texture, hughes2018robotic} and soft actuators \citep{polygerinos2017soft}. Examples of such signals are shown in Figure \ref{fig:touchdata}.

This diversity in sensing is captured by policy learning, with our reviewed works leveraging all mentioned forms of sensing, across different products and robots. These works all demonstrate an intuitive result, which is that conditioning robot policies on force sensing enables robot skills that are otherwise limited, inferior, and/or impossible without touch sensing, such as pouring precise volumes from a cup \citep{seehear, lee_making_2019}, inserting pegs into tight-tolerance holes \citep{msbot, visk}, or grasping fragile and deformable objects \citep{dexforce, factr, jaf}.

\begin{figure*}[!th]
\centering
\captionof{figure}{Touch sensing can be represented across fine-grained fingers to the whole robot arm as forces.}
\begin{subfigure}[b]{0.42\linewidth}
\centering
    \includegraphics[width=0.9\linewidth]{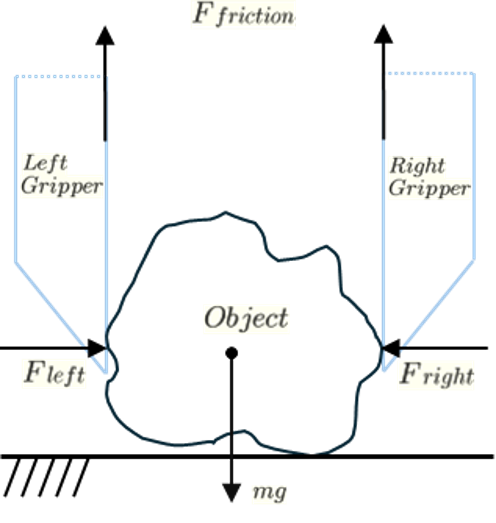}
    \caption{Free body diagram of gripper-object interaction}
    \label{fig:grasp_force}
\end{subfigure}
\hfill
\begin{subfigure}[b]{0.5\linewidth}
\centering
    \includegraphics[width=0.9\linewidth]{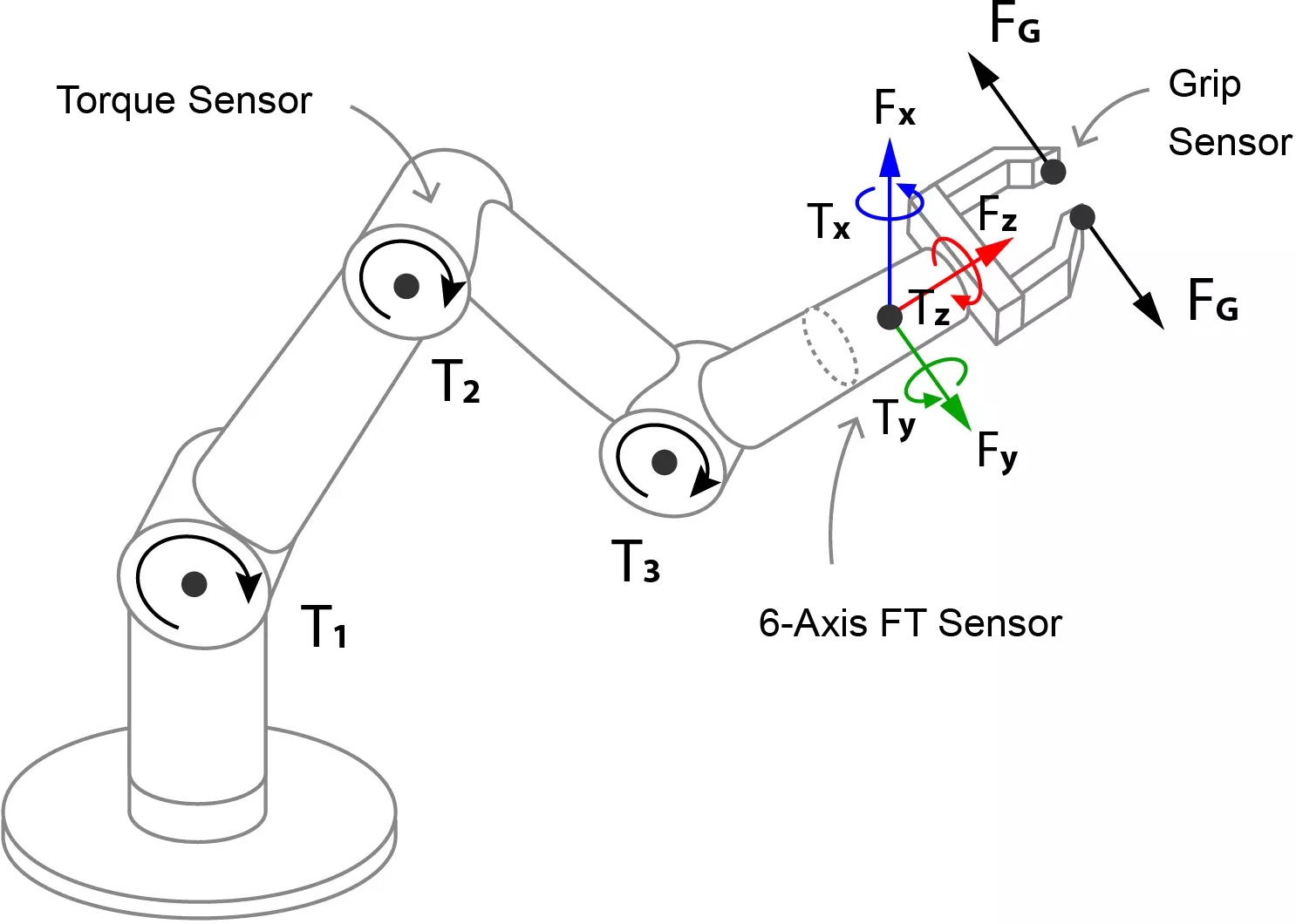}
    \caption{Common force or touch sensing methods on robot arms include joint torques, wrist force/torque (F/T) sensors, and end effector gripper sensors. Diagram from \citep{force_diagram}.}
    \label{fig:force_diagram}
\end{subfigure}
    \label{fig:forces}
\end{figure*}

During grasping, a robot end effector applies an external force on the object, as expressed in Fig. \ref{fig:grasp_force}. Assuming sufficient friction, at least two points of contact are needed for to create sufficient constraints on the object. Note, that the drawing shows the end-result of grasping, but contact with the objects and the different fingers of the gripper or hand almost never happen at the same time (see also Figure \ref{fig:touchdata}), which creates a strong motivation for tactile sensing \cite{patel2017improving} to minimize disturbance of an object as individual fingers are placed.  

Once the robot moves, this external wrench propagates through the arm, generating internal wrenches that must be accounted for in the robot's control. In this view, touch is represented as the forces and torques transmitted through the end effector to the robot, which can be interpreted either as a six-dimensional wrench at the wrist or, by solving the full inverse dynamics, as joint torques across the robot’s degrees of freedom, depicted in Fig. \ref{fig:force_diagram}. Beyond grasping, this representation also applies to pushing or pulling an object with the robot end-effector. 


\subsection{Foundation Models}\label{sec:large_data}
Large-scale robotic datasets \citep{openx, droid} have enabled the emergence of generalist, end-to-end robot foundation models \citep{openvla, rt1, visualcortex, octo,  ecot, tinyvla} which typically combine a vision-language model with a behavior cloning architecture \citep{bet, act, vqbet, dp} to generate robot policies from a larger representation space. However, these robot foundation models are pre-trained on limited modalities: vision, language, and robot joint and/or end effector data. This includes the most recent Gemini Robotics \citep{gemini_robotics}, which has recorded 2000-5000 episodes per task across six tasks and over a time-span of 12 months, but does not include force. 

Recently, we have seen an glut of smaller robot policy models which do capture and condition on tactile feedback, positioned at varying levels of generality and task and problem coverage. Such works have inherited the combinatorial, heterogeneous nature of physical sensing, with each contribution often proposing a unique ensemble of solutions for tactile policy learning. In this review we examine select key questions in this space: 1) how should we collect touch data in robot motion, 2) how should we express robot actions conditioned on touch data, and 3) how do we represent this data in robot policy learning?

\section{Review Overview} \label{sec:review}
\vspace{-6pt}
In this section, we describe the review structure of 25 works, which learn tactile robot policies using transformer or diffusion models: \citep{ablett_multimodal_force_control_switching_matching_2024, dipcom, dexforce, msbot, he_foar_2024, acp, huang_3d-vitac_3dvitac_2024, fuse, kobayashi_alpha-biact_2024, lenz_analysing_2024, maniwav, liu_forcemimic_force_control_2024, factr, mejia_hearing_2024, noseworthy_forge_2024, visk, romero_eyesight_2024, poco, wu_tacdiffusion_2024, jaf, xu_unit_2024, yu_mimictouch_2024, t3, zhou_admittance_2024, seehear}. Unlike previous work on end-to-end learning of force-based policies \citep{buchli2011learning,abu2020variable,zhang2021learning,park2023lstm}, transformer and diffusion-based methods inherit the favorable scaling properties of large language models, making them suitable for training foundation models. 

\begin{figure}[!htb]
    \centering
    \resizebox{\textwidth}{!}{\input{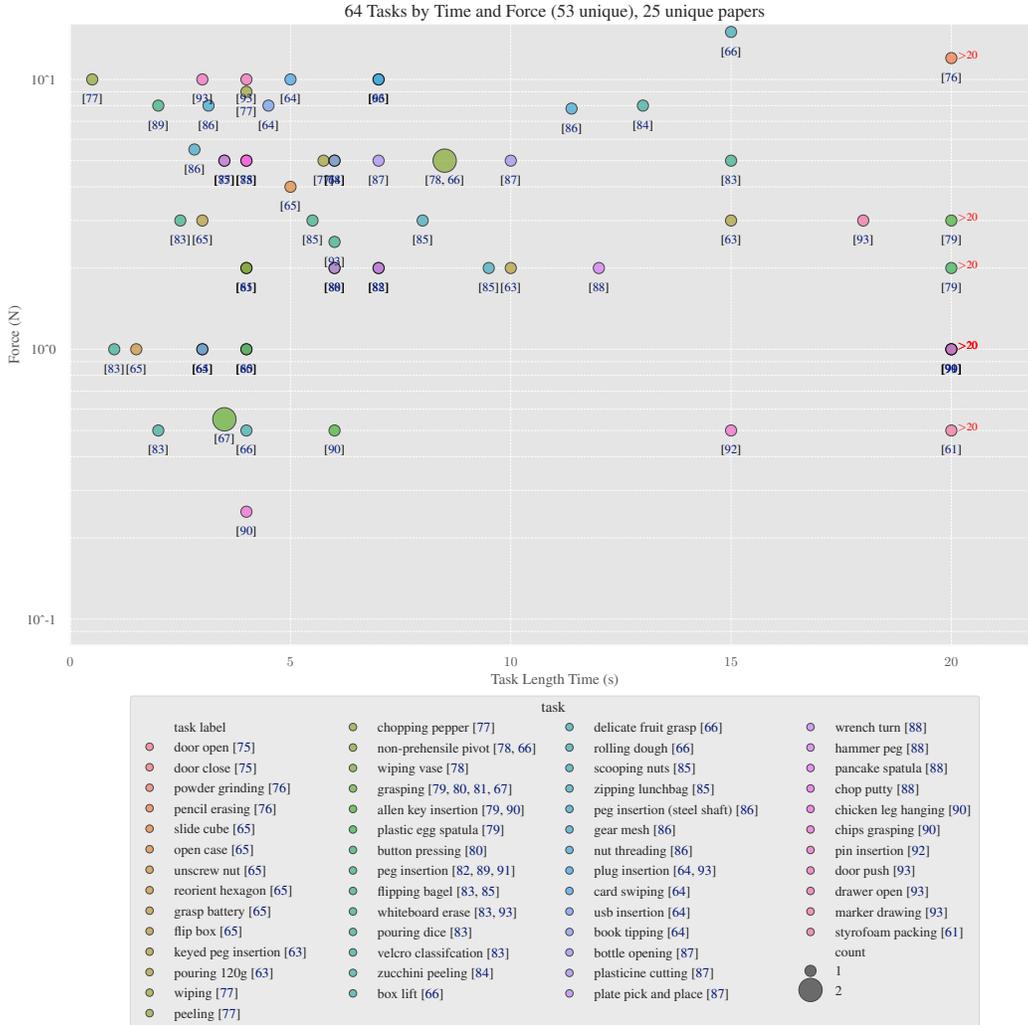}}
    \caption{On the same force-time axes, we describe the 64 tasks learned by the 25 papers, with 53 unique tasks.}
    \label{fig:tasks}
\end{figure}

From these papers, we identify 64 manipulation experiments, corresponding to 59 distinct policies trained and 53 unique tasks (Figure \ref{fig:tasks}). That is, there is neither a consistent challenge application, except perhaps ``peg-in-hole'' for which five papers provide benchmarks, nor a policy that is capable of dealing with more than a handful of tasks at once.

\begin{figure}[!htb]
    \centering
    \resizebox{\textwidth}{!}{\input{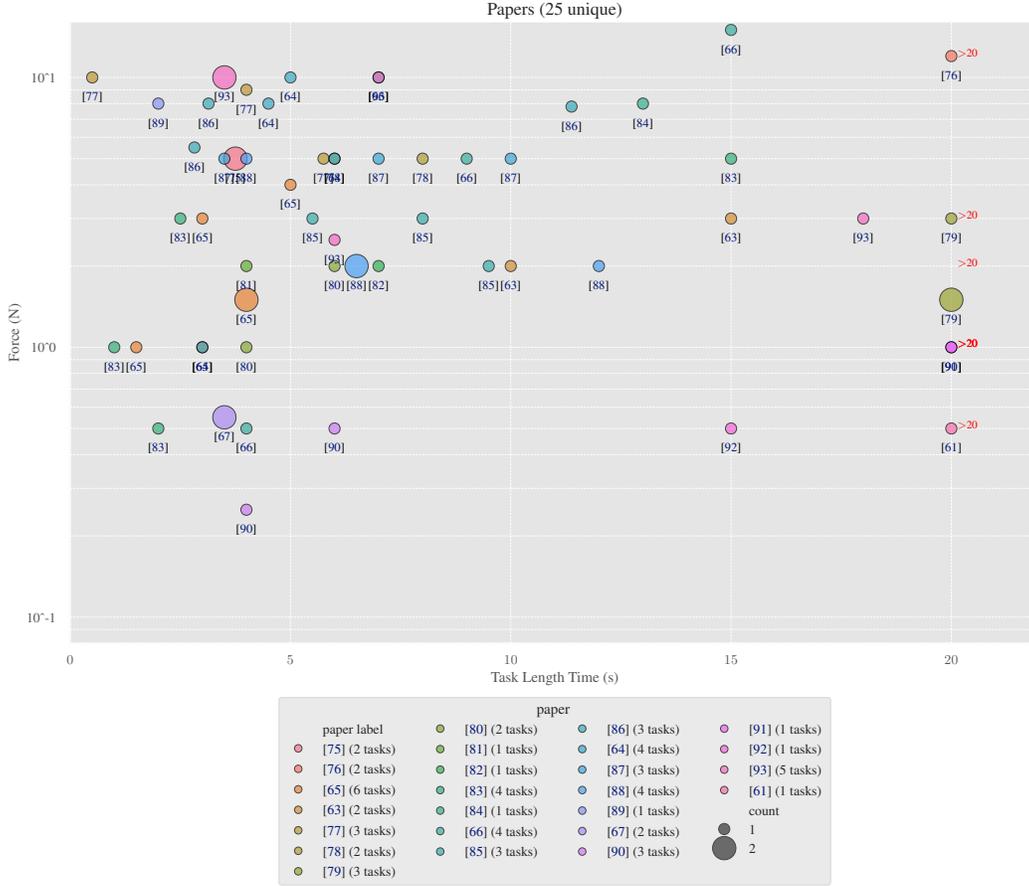}}
    \caption{We plot force magnitude against task length time for 64 tasks (of which 53 are unique) across 25 papers implementing tactile robot policies.}
    \label{fig:papers}
\end{figure}

We employ multiple lenses with which to analyze these works and their respective tasks, the first of which is to plot the approximate force magnitude from 0.1 to 10N against task length time from 0.1 to 20s (``makespan''), categorized by paper in Fig. \ref{fig:papers} and by specific task in Fig. \ref{fig:tasks}. Within these works, we specifically explore their data collection methods in Sec. \ref{sec:data_collection}, action spaces in Sec. \ref{sec:action_space}, and representation learning methods in Sec. \ref{sec:policy_learning}.

As only seven of the reviewed works (28\%) provide force magnitudes for their learned tasks, we have estimated the order of magnitude for the other works. We represent force magnitude in logarithmic bins between 0.1N and 1N (delicate force), 1N to 10N (typical operational force), and greater than 10N (high force). We do an additional rough categorization within these bins based on the specific task. Although some works do not provide measured task length times, we are able to estimate task length from videos provided on project websites or presentations in such cases. Separating tasks between short and long duration ($\leq$ or $>$ than five seconds), we find that 4 tasks (6\%) are short ($\leq$5s) and apply high forces ($\geq$10N), 14 (22\%) are short and apply typical forces (1N to 10N), 11 (17\%) are short and apply delicate forces (0.1N to 1N), 6 are long ($>$5s) and apply delicate forces, 25 (40\%) are long and apply typical forces, and  4 (6\%) are long and apply high forces. 

It is possible to construct similar taxonomies categorized by touch sensor type, data collection method, policy learning architecture, dataset size, or policy action space. However, due to the high visual density of such resulting plots, in future sections we narrow our focus on the distribution of unique papers across a single category (e.g. what is the distribution of touch sensor type across the 25 works). 


We provide a full reference table for the 25 works and corresponding 64 tasks containing information, when available, on approximate force magnitude, task length time, general touch sensor type, specific touch sensor, policy action frequency, dataset size (per-task), action space of policy-generated actions, data collection method, low-level robot controller, policy learning architecture, and miscellaneous notes (typically relating to data representation) via this \hyperlink{https://docs.google.com/spreadsheets/d/1ipFKCbHVZmLyGin1R6eO5_re84M3OOqHOtpzRIr6dXI/edit?usp=sharing}{online spreadsheet} (\hyperlink{https://docs.google.com/spreadsheets/d/1ipFKCbHVZmLyGin1R6eO5_re84M3OOqHOtpzRIr6dXI/edit?usp=sharing}{link}).

\begin{figure}[!htb]
    \centering
    \resizebox{\textwidth}{!}{\input{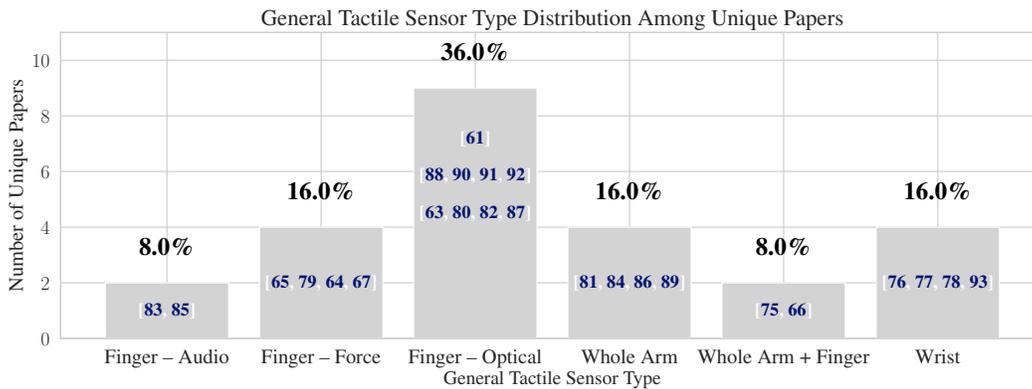}}

    \captionof{figure}{Across the reviewed papers, we categorize touch or tactile sensors across six categories: audio, force, or optical (visuotactile) sensing at the end effector ``fingers," joint torque sensing along the whole robot arm, combined sensing from the end effector and joint torques, and wrist F/T sensing.}
    \label{fig:sensor_dist}
\end{figure}


\section{Data Collection}\label{sec:data_collection}
A fundamental challenge in policy learning is data collection of high-quality robot trajectories, which is predominantly accomplished by a human ``demonstrating" how to do a specific task. Training a robot foundation model, e.g. a very large robot policy capable of many tasks across diverse environments and configurations, requires a proportionally very large amount of this high-quality robot data, typically characterized as a scaling law in machine learning \citep{brown2020gpt3, lin2025datascalinglawsimitation}. Tactile robot policy learning particularly adds difficulty to this scaling law. In a field where sensing varies significantly across platforms, data collection methods necessarily also vary, and thus it is difficult to amass the requisite amount of robot data for a tactile robot foundation model. In this section we discuss this first problem of extracting touch sensing for robot motion data, examining how reviewed works design data collection methods to capture touch sensing for their specific tasks and reduce the human-robot embodiment gap in demonstrations.

\subsection{Sensor modalities}
The first single-category lens we look at is general touch sensor type, shown in Fig. \ref{fig:sensor_dist}. We categorize touch sensing across six categories: audio, force, or optical (visuotactile) sensing at the end effector ``fingers", joint torque sensing from the robot arm joints (``whole arm"), combined sensing from the end effector and joint torques, and wrist F/T sensing. Visuotactile sensors at the finger are entirely constituted of GelSight-type \citep{yuan2017gelsight} sensors, accounting for the plurality of sensing (36\% of papers). The GelSight-type sensors measure touch by observing the deformation of a flexible polymer using a camera.
Sensing is otherwise diverse, with the other methods constituting two (finger audio and combined arm and finger sensing) to four (finger force, whole arm, and wrist F/T) papers each. 

In total, 14 unique sensor products are used, including the single category of GelSight-family sensors \citep{msbot, fuse, lenz_analysing_2024, romero_eyesight_2024, poco, xu_unit_2024, yu_mimictouch_2024, t3, seehear}. Joint torque sensing across whole robot arms is accomplished via off-the-shelf sensing from the Franka Panda robot arm \citep{ablett_multimodal_force_control_switching_matching_2024, factr, noseworthy_forge_2024, wu_tacdiffusion_2024} or Flexiv Rizon arm \citep{liu_forcemimic_force_control_2024}, or from motor current sensing on custom robot arms \citep{kobayashi_alpha-biact_2024}. Wrist F/T sensing can be done also with the Franka Panda arm or with three other dedicated wrist sensors: the UR5E sensor \citep{dipcom}, OptoForce sensor \citep{he_foar_2024, zhou_admittance_2024}, and the ATI Mini 45 sensor \citep{acp}. For finger audio sensing both surface microphones \citep{maniwav, mejia_hearing_2024} and normal microphones \citep{fuse} are used. For finger force sensing, motor current corresponding to contact normal force \citep{factr, jaf}, CoinFT sensors \citep{dexforce}, and uncalibrated force-like quantities like magnetic-field sensing \citep{visk}, and pressure-sensing \citep{huang_3d-vitac_3dvitac_2024} are used.

\begin{figure}[!htb]
    \centering
    \resizebox{\textwidth}{!}{\input{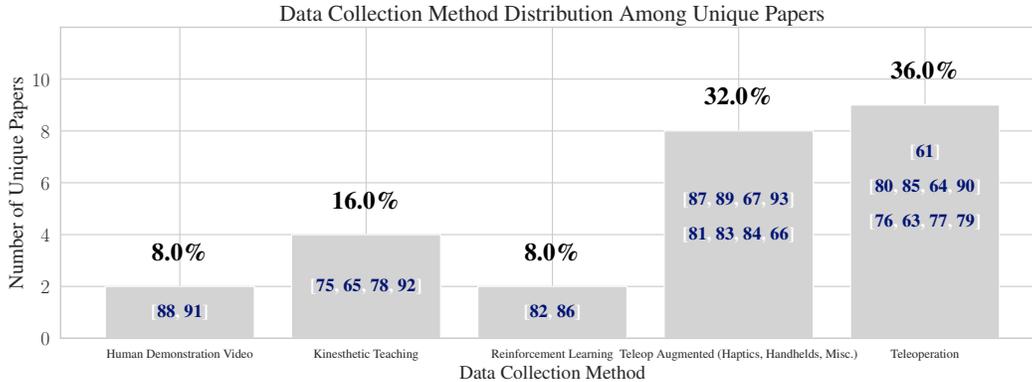}}
    \captionof{figure}{We categorize data collection into five methods: 1) human video demonstration methods, 2) kinesthetic teaching, which entails directly teaching the physical, human forces to perform a task, 3) reinforcement learning methods which do not require demonstration data, 4) augmented methods of teleoperation including haptic feedback for the operator, handheld grippers detached from a robot arm, and force-matching leader-follower arm systems, typically with ALOHA-like systems \citep{act}, and 5) teleoperation.}
    \label{fig:data_collect_dist}
\end{figure}

\subsection{Teleoperation}


\emph{Teleoperation}, whereby an expert operator uses a joystick, potentially with a VR headset \citep{dipcom, fuse, mejia_hearing_2024, visk, poco}, keyboard and other miscellaneous computer devices \citep{msbot, seehear}, or via a smaller robot arm in a leader-follower setup \citep{huang_3d-vitac_3dvitac_2024, xu_unit_2024}, to control the robot, accounts for the plurality of works (36\%), Figure \ref{fig:data_collect_dist}.


\subsection{Kinesthetic Teaching}
%
%
%
The embodiment gap between human operators and robotic systems arises because robots experience forces differently from humans, both in magnitude and in the way forces are applied and sensed. Kinesthetic teaching mitigates this discrepancy, allowing operators to directly impart forces onto the robot by physically guiding it through a task. 
Kinesthetic teaching entails directly teaching the physical, human forces to perform a task \citep{ablett_multimodal_force_control_switching_matching_2024, dexforce, acp, yu_mimictouch_2024, t3}. Hou et al. place two UR5E robots in free-drive mode with gravity compensation, allowing them to move freely while being guided by a human operator, a common approach in kinesthetic teaching \citep{acp}. The human demonstrator then grabs specially designed handles mounted on the robots' wrists, reminiscent of bimanual exoskeleton control, and demonstrates two high force magnitude tasks: cleaning a vase with sponge end-effectors and pivoting an object about a rigid surface with a rod end-effector. Ablett et al., in comparison, demonstrate more typical forces and move a Franka Panda arm to open a cabinet door, recording external forces via the arm's joint torques \citep{ablett_multimodal_force_control_switching_matching_2024}. Finally, Zhao et al. use kinesthetic teaching to do precise, low-force capacitor insertion on a printed circuit board, recording forces indirectly with the GelSight Wedge sensor \citep{t3}. 

In these works, the operator directly perceives forces through the robot’s arm, and the robot’s sensors, ranging from finger, wrist, and whole arm sensing, capture the imparted human forces, ranging from low, typical, and high forces. This method provides rich force interaction data representative of the nuanced compliance strategies at play in tasks that involve external contact forces. 
Unlike teleoperation, which relies on indirect control interfaces, kinesthetic teaching enables the robot to record internal and external wrenches (i.e., forces and torques applied by the human through both the robot and the environment) in a manner more representative of real-world forceful interactions.

Kinesthetic teaching can also be accomplished by recording forces from a human demonstrator using finger-mounted sensors. Two works explore directly attaching a GelSight Mini \citep{yu_mimictouch_2024} or CoinFT (finger F/T sensor) \citep{dexforce} to a human demonstrator's hand as they perform pinch grasps and single-finger motions, effectively demonstrating the forces to be emulated on a parallel-jaw robot gripper. This design enables direct and intuitive demonstration of contact-rich manipulation skills, such as precise peg insertion \citep{yu_mimictouch_2024}, object reorientation, articulated grasping of earphone cases and enclosed batteries, and non-prehensile sliding \citep{dexforce}. 

Several issues pervade kinesthetic teaching, however: it is physically demanding, potentially time-intensive, and risks damaging fragile robotic sensors or actuators if mishandled, thus requiring skilled human demonstrators, of which there is often scarce supply.

\subsection{Augmented, Bilateral Teleoperation}
A middle ground between kinesthetic teaching and teleoperation is bilateral control, or augmented teleoperation, which incorporates haptic or force feedback from the robot in teleoperation. In these setups, forces sensed by the robot are mirrored back to the human operator, enabling closed-loop interaction, using force-matching leader-follower arm systems \citep{kobayashi_alpha-biact_2024, factr}, haptic feedback joystick devices \citep{zhou_admittance_2024, romero_eyesight_2024}, or with hand-held robot grippers equipped with touch sensing intended to emulate a robot end-effector \citep{maniwav, liu_forcemimic_force_control_2024}. Some works additionally mix robot demonstration data with human demonstration data \citep{poco, yu_mimictouch_2024}, or teleoperated robot data with handheld gripper data \citep{maniwav}. In total, these alternative methods for teleoperation account for 56\% of the reviewed works, though they come at the cost of additional required expertise and system design. 

The magnitude of force feedback from bilateral teleoperation can be scaled to help operators develop an intuitive feel for the task dynamics without exerting the true, full forces required to complete the task. Compared to kinesthetic teaching, bilateral control reduces the physical burden on the human while maintaining some degree of force awareness. However, bilateral control systems are highly engineered and often task-specific. The feedback provided to the operator is not a direct measurement of either human-applied forces or robot-experienced forces, but rather a processed signal reflecting robot interaction forces. 

Designing effective feedback mechanisms is nontrivial. Researchers have explored a variety of techniques, ranging from vibration-based hand feedback \citep{zhou_admittance_2024} to leader-follower robotic arms that attempt joint torque transfer without unduly burdening the operator \citep{factr, kobayashi_alpha-biact_2024}. Xue et al. propose force-field visualizations, mapping interaction forces into a graphical display rather than physical feedback \citep{xue2025reactivediffusionpolicyslowfast}. In constructing such systems, these works often near the complexity of kinesthetic teaching, demanding substantial hardware equipment and expertise, making data collection still expensive and difficult to scale.

Handheld robot grippers are promising alternatives that simplify the data collection process while preserving force fidelity. This method removes the robot arm from the touch-sensing feedback loop entirely, focusing on force interactions at the end-effector. The assumption undergirding these devices is that force sensing at the wrist and gripper is the relevant signal for many manipulation tasks. Given this, researchers have developed portable force-sensing grippers equipped with F/T sensors and wrist-mounted cameras for direct data collection by human demonstrators \citep{liu_forcemimic_force_control_2024, zou2025fewshotsim2realbasedhigh, hagenow_versatile_vdi_2024}.

Handheld grippers offer several key advantages: 1) direct force measurement; unlike bilateral control, these devices capture human-applied forces at the end effector without additional signal processing, 2) reduced complexity and cost; they eliminate the need for full robotic systems, lowering the barrier to collecting high-quality force-interaction data, and directly related to the prior advantage, 3) improved scalability; these tools are easier to use, vastly more portable, and require less expertise than kinesthetic teaching or leader-follower methods.

Recent work has also explored diverse designs for handheld force-sensing grippers. Liu et al. integrates contact microphones at the fingertips of the handheld UMI gripper \citep{chi2024universalmanipulationinterfaceinthewild} to approximate force feedback via audio signals \citep{maniwav}. This approach enables highly sensitive tactile tasks, such as distinguishing surface textures (e.g., hook and loop tape surface identification), by learning the acoustic properties of frictional contact. Though these compact, specialized touch-sensing devices present practical and scalable alternatives for data collection in tactile robot policy learning, by disregarding the robot arm in demonstrations, future work should take care in performing high-force magnitude tasks. When learning to generate correspondingly forceful actions at the end-effector, dangerous joint space trajectories may be enacted by the robot.

\begin{figure*}[h]
    \centering
    \resizebox{0.8\textwidth}{!}{\input{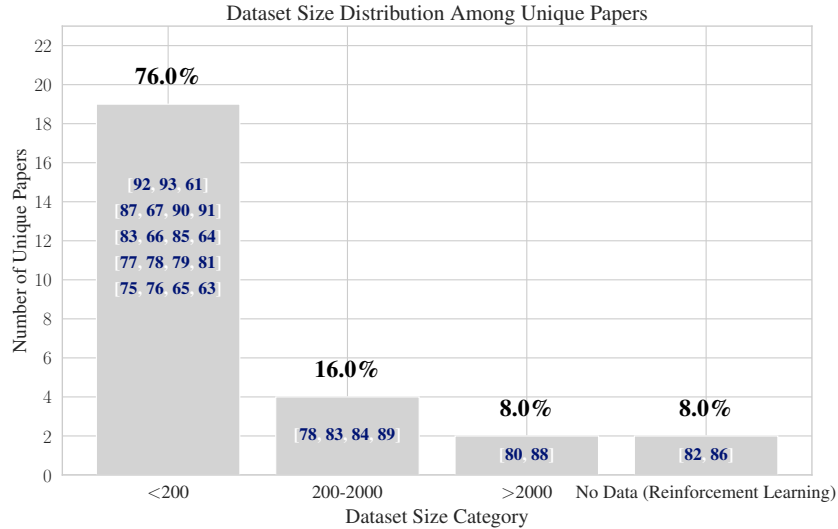}}
    \captionof{figure}{The large majority of papers (76\%) of papers train their policies on under 200 demonstrations. The exceptions collect large amounts of data either in simulation (Drake) \citep{poco} or via high-manpower data collection efforts \citep{fuse}.}
    \label{fig:dataset_dist}
\end{figure*}



\subsection{Data requirements for forceful policies} \label{sec:scaling}
Of the reviewed papers, the overwhelming majority (76\%) train on fewer than 200 demonstrations, shown in Fig. \ref{fig:dataset_dist}. This reflects the broader challenge of collecting large-scale, high-quality force interaction datasets across heterogeneous robot platforms and suggests that, as no clearly superior sensing or data collection method has emerged, it may be premature to scale data collection. However, two outlier works train policies on datasets exceeding 2,000 \citep{fuse, poco} episodes. 

The first such work from Jones et al. trains a policy, FuSe, on 26,866 tactile trajectories collected en-masse via VR headset teleoperation on Widow X robots outfitted with GelSight DIGIT sensors and microphone audio sensing at the fingers, in addition to language instruction, wrist camera vision, and third person camera vision \citep{fuse}. This dataset is by an order of magnitude the largest real-world robot dataset with touch sensing. The resulting trained policy is able to distinguish textural and auditory features, generalizing to grasping new objects with novel and varying touch properties. However, while the large dataset of multimodal data enabled the trained policy to semantically reason about and classify objects based on their tactile properties, the policy learned primarily to grasp objects and press buttons (two distinct tasks) conditioned on this knowledge. With teleoperation, scaling up to skillful and nuanced manipulation tasks presents significant challenges, requiring proportionally greater manpower, time, and expertise.

Wang et al. propose an alternative approach to scaling data collection without human data collection altogether, instead training policies primarily on 50,000 contact-rich, continuous tool-usage skills obtained in simulation (Drake), leveraging simulated GelSight sensing \citep{poco, wang2024robotfleetlearningpolicy}. A policy trained on simulation data combined with 400 real robot demonstrations exhibits better generalization to novel objects, distractors, and configurations as a result of domain and configuration randomization deployed in simulation. However, the simulated tasks are significantly abstracted; for example, in the hammer usage task, the demonstrated dynamics are not representative of practical usage. Simulation remains an underexplored domain for learning high-quality tactile robot policies, but perhaps the underlying condition of a simulator capable of accurately modeling dynamic, frictional, and large-magnitude forces has not been met yet. 

We argue that it is never premature to scale data collection, if provided ample means to do so. Such efforts yield insights into model training and data representation distinct from works which are devoted to data collection method design and train on smaller datasets. 
Also, even if additional modalities are required later, such data might still be relevant to initialize a model with curriculum learning.

\begin{figure*}[h]
    \centering
    \resizebox{0.66\textwidth}{!}{\input{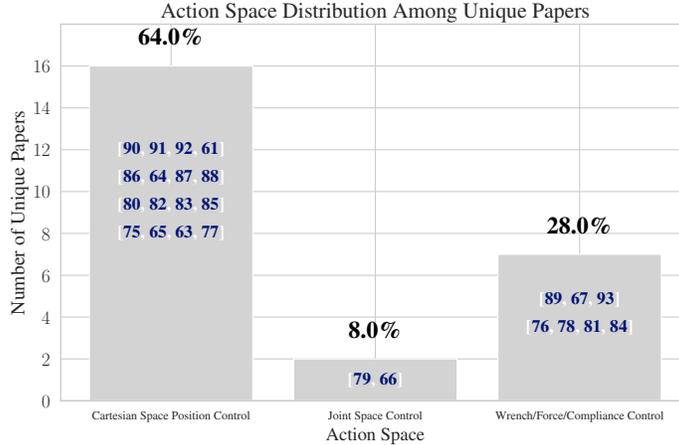}}
    \captionof{figure}{
    A large (64\%) majority of learned policies output robot actions in Cartesian space position control. Outside of this method of control, low level control is split across joint space control and various forms of force control.}
    \label{fig:action_dist}
\end{figure*}

\section{Action Space}\label{sec:action_space}
While in the previous section we described a wide array of data collection methods, due to the primary underlying method being teleoperation in Cartesian space, a similarly large majority of learned policies (64\%) output robot actions in Cartesian position space for a low-level robot position controller. Low level control is otherwise split across joint space control and various forms of force control--admittance \citep{acp, zhou_admittance_2024}, impedance \citep{ablett_multimodal_force_control_switching_matching_2024, dexforce, noseworthy_forge_2024, wu_tacdiffusion_2024}, compliance control \citep{dipcom}, and other custom control schemes \citep{kobayashi_alpha-biact_2024, liu_forcemimic_force_control_2024}). In this section we discuss how low-level force control can be formulated from human demonstrations, what benefits it yields, and whether it is necessary at all.

It is perhaps misleading to dichotomize policies by low-level control. After all, if a position-control policy learns robot motion conditioned on force feedback, then it itself is a model-free, implicit force-position controller, albeit highly specialized for the learned task \citep{factr, he_foar_2024}. With the concrete disadvantages of requiring more complex  control schemes and potentially processing demonstration data to force-controllable inputs, explicit low-level force control's concrete advantages are then: 1) performance, as such controllers run at high frequencies that enable reactivity and consistency beyond human capability, which are often further gimped by wide embodiment gaps in demonstration, 2) interpretability, in that force controllers accept low-dimension motion objectives or parameters that can be explicitly commanded, anticipated, and intuitively understood (e.g. maintaining a commanded scalar compliance parameter), and 3) dimension reduction in policy learning, in that learning motion parameters rather than direct robot motion offloads the complexity of force control from the policy to the controller. 

\subsection{Explicit Force Control}
It is still possible to generate force controllable actions even if robot demonstrations operate in Cartesian space. When such demonstrations are collected with force data, one can formulate and reconstruct force-controllable actions to be trained on. Revisiting Hou et al. \citep{acp}, their trained Adaptive Compliance Policy performs the vase-wiping task and quasistatic flipping task by mapping human-demonstrated forces to high-frequency (500 Hz) admittance controller inputs. To briefly revisit \ref{sec:bg_force_control}, such a force control scheme governs robot motion as a mass-spring-damper system, taking control inputs of a virtual target pose and stiffness matrix in response to external forces. 

To achieve this, Hou et al. design a post-processing method to reshape wrist force sensing (ATI Mini F/T wrist sensor) and end-effector position data from kinesthetic teaching demonstrations to admittance controller inputs, in order to train a policy able to command variable compliance across different contact modes and disturbances. Hou et al. formulate a post-facto stiffness matrix control input with the heuristic of allowing low stiffness (high compliance) in the direction of the force feedback and high stiffness otherwise. This low stiffness value is scaled by sensed force magnitude. Then, they project a virtual target position from position data in demonstrations along the sensed wrist forces and scaled by the computed stiffness. Additionally, a 1-second moving average filter is applied to demonstrated wrist wrenches to generate future-contact-informed stiffness inputs, which subsequently produce smooth, contact-engaging virtual target trajectories. Finally, the tactile robot policy is trained to predict a virtual target and stiffness value, in addition to true end-effector pose. As a result, the policy maintains appropriate compliance throughout unseen perturbations (jostling) and geometries (vases and objects to pivot). Compared to ablated policies which do not learn variable compliance and use a uniformly high- or low-stiffness controller, closer to position control, absolute success rate drops by 81\% for the same tasks.

Other works follow a similar implementation of reconstructing force-informed trajectories post-demonstration. Chen et al. \citep{dexforce} use fingertip F/T sensing and Ablett et al. map low-dimension deformation signals from a pressure-based finger tactile sensor to forces \citep{ablett_multimodal_force_control_switching_matching_2024}. These works utilize finger sensing rather than wrist F/T data in order to decouple wrist wrenches, which are inextricable from human-applied wrenches if demonstrated with kinesthetic teaching, from the precise forces experienced at the fingers, before generating virtual targets with tuned stiffness components for impedance control. These approaches require careful model design and tuning of forceful action representation, but enable large success rate improvements and robust compliant behaviors for tasks like reorienting objects, opening doors, and manipulating other kinds of articulated objects compared to ablated methods without force input and force-informed virtual targets. 

Zhou et al. also leverage admittance control but directly predict future contact forces while adjusting to real-time contact forces, rather than predicting stiffness control parameters to generate future force-informed virtual targets \citep{zhou_admittance_2024}. The trained tactile policy performs tasks such as grasping, cabinet and door opening, dry-erase board drawing and erasing and demonstrate improved success, a 17\% reduction in task completion time over a policy trained without force feedback, and a 26\% reduction in task completion time over teleoperated methods without force control.

Control schemes such as dually tracking orthogonal pose and force targets \citep{liu_forcemimic_force_control_2024}, predicting real, non-virtual poses and stiffness parameters for an underlying compliance controller to resolve into target joint positions \citep{dipcom}, or simply commanding to a grasping force, rather than gripper position or closure \citep{jaf} are also employed. Trained policies from \citep{liu_forcemimic_force_control_2024, dipcom} demonstrate task performance on mortar-and-pestle grinding and zucchini peeling close to human expert time efficiency (1.3x) and three times faster than teleoperated methods (4.5x).  

Noseworthy et al. \citep{noseworthy_forge_2024} generate actions as Cartesian space virtual targets for low-level impedance control, but do not train on stiffness or force targets. Instead, the trained policy learns the inherent impedance control law from contact forces provided in the observation. In order to generalize across a range of contact forces, they simulate impedance control and Franka Panda robot dynamics with randomized scaling and damping parameters, spanning commandable forces between 6 and 20N. Ablated policies trained without force input and thus doing solely position control were less successful, took longer to complete, and exerted larger ground-truth forces on the same tasks.

Wu et al. train a policy to directly output high-frequency target external wrenches between 50 to 500 Hz, which are initially collected from pre-programmed behavior-tree guided peg insertion demonstrations \citep{wu_tacdiffusion_2024}. They additionally implement dynamic filtering to interpolate from the policy action generation frequency to a low-level impedance controller at 1000 Hz. These trained policies execute precise ($<$0.5mm tolerance) peg insertion, on average, in under two seconds and above 90\% success rate, where similar tactile robot policies generating position control actions to do less-precise peg insertion require typically at least double the time \citep{msbot, lenz_analysing_2024, yu_mimictouch_2024}.

\begin{figure*}[h]
    \centering
    \resizebox{\textwidth}{!}{\input{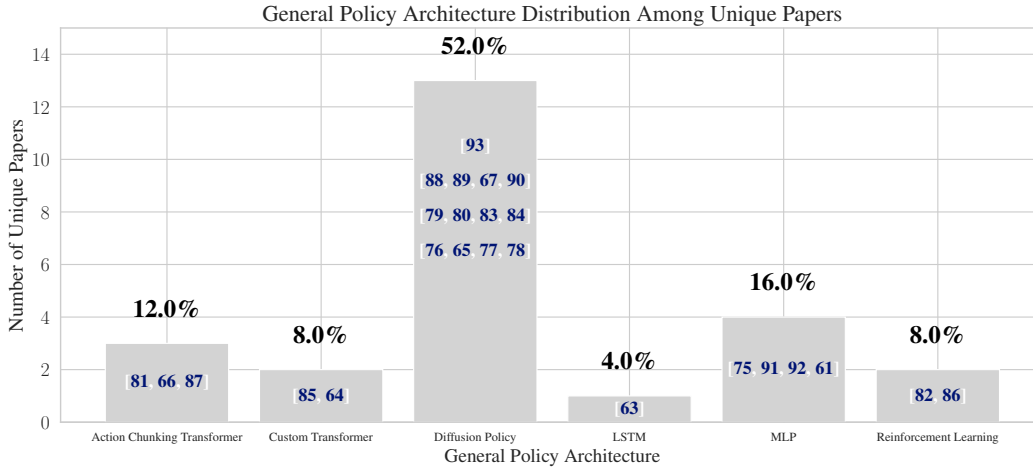}}

    \captionof{figure}{Looking at the behavior or policy learning architectures utilized by the reviewed papers, the majority (52\%) train their policies with a diffusion architecture backbone \citep{dp}. Five papers (20\%) use transformer architectures, with three of those using the action-chunking transformer introduced by \citep{act}. Multilayer perceptron architectures (16\%) are still relevant for action generation due to low quantities of data.}
    \label{fig:policy_dist}
\end{figure*}

\section{Policy Learning}\label{sec:policy_learning}
As policy learning is the natural bottleneck between sensory inputs and robot actions, the reviewed works implement various approaches to reduce dimensionality in tactile robot policies. In this section we discuss selection of behavior cloning architecture in policy learning and the differences between representing force data and visuotactile data for policy learning. 

\subsection{Behavior Cloning}
The majority of reviewed works (52\%) train their policies with a diffusion architecture backbone \citep{dp}, shown in Fig. \ref{fig:policy_dist}. Five papers (20\%) use transformer architectures, with three of those using the action-chunking transformer (ACT) \citep{act}. Multilayer perceptron architectures (MLP) are still leveraged by some works (16\%) as they can learn behavior cloning with low quantities of data. 

Diffusion policies have gained traction primarily due to their ease of training and their ability to balance sample efficiency with the capacity to model complex, multimodal robot behaviors by capturing the stochastic nature of human demonstrations. Unlike transformers, which often require large datasets to generalize effectively, or small MLPs, which struggle with intricate tasks and mode collapse, diffusion policies offer a middle ground. Diffusion does suffer some drawbacks, such as overfitting to absolute robot states and reducing generalization to other spatial configurations. Removing robot states from the observation and instead leveraging relative, instead of absolute, position actions improves this issue \citep{octo, fuse}, allowing tactile diffusion policies to learn from more relevant sensing.

ACT \citep{act} address a different challenge in policy learning: long-horizon reasoning and efficient action execution. By structuring actions into temporally coherent chunks, ACT reduce the burden of high-frequency action prediction while maintaining smooth and stable control. This is especially useful in tactile robot tasks requiring extended, coordinated motion sequences. The chunking mechanism allows the transformer to learn meaningful action segments, effectively bridging the gap between low-level motor commands and high-level task objectives.

While behavior cloning architectures have converged toward effective paradigms, another challenging aspect of policy learning lies in tactile representation learning, as in capturing salient tactile features in learnable, lower-dimension features.

\subsection{Visuotactile Representation Learning} 
Of the 14 unique sensors employed in the reviewed works, tactile representation learning research largely focuses on one: GelSight-type sensors which capture high-resolution surface deformation images, making them powerful tools for visuotactile learning.

For instance, Jones et al. \citep{fuse} fine-tune a TVL encoder \citep{tvl}. pretrained on 44,000 vision-touch-language samples, to represent visuotactile sensing from a GelSight DIGIT sensor. This encoder, built on a vision transformer (ViT) \citep{vit}, integrates annotated vision, language, and touch data from various GelSight sensors, mapping visuotactile data to semantic features like texture, force, and object category. The TVL encoder leverages vision-language pretraining (VLP), where large-scale multimodal datasets enhance representation learning. By aligning visuotactile signals with semantic and visual concepts, the latent representations of tactile data enable a richer understanding of contact interactions. In comparison, Xu et al. design a representation which can learn from a single object and sensor to reconstruct visual deformation of novel objects and from other GelSight-type sensors \citep{xu_unit_2024}. Zhao et al. \citep{t3} propose a tactile encoder for directly learning downstream tasks such as classification, force estimation, and pose estimation from representations of visuotactile data from different GelSight-type sensors. 

Despite the inherent challenges of working with visuotactile data from GelSight-type sensing, such sensors are more accessible than force-sensing methods, which require specific robot arms such as the Franka Panda or comparatively expensive wrist F/T sensors. As a result, research ecosystem surrounding these sensors continues to grow, continually improving learning from visuotactile inputs.

Pattabiraman et al. leverage a single magnometer-based finger sensor (AnySkin \citep{anyskin}), which provides neither optical deformation nor force data \citep{visk}. However, the measured sensor produces data similar in dimension to force data, as it is equipped with five 3-axis magnetometers, totalling a 15-dimensional sensor reading. This data allows easier representation learning, as low-dimensional signals across objects ease estimation of contact events and contact magnitude. As such, the sensor data is encoded with just two fully-connected layers. The sensor provides uncalibrated force-like quantities (magnetic force fields) that capture contact shear and normal forces more directly than visual deformation, which enables learning continuous, precise tasks such as credit card swiping, tipping over and grasping a book off a shelf, and plug insertion from the lightweight data representation. Huang et al. learn from  pressure sensing at the finger, which also provides neither force nor optical data \citep{huang_3d-vitac_3dvitac_2024}. However, they take an alternative approach and project tactile sensing and depth camera vision to a shared 3D, point-cloud representation to improve bi-manual, in-hand manipulation.

\subsection{Force Representation Learning}
Force representation learning is less explored and oftentimes more straightforward. Force data is low-dimensional and explicitly, causally linked to motion. Unlike image data, force measurements are interpretable in their raw forms, can be encoded often directly without modification \citep{ablett_multimodal_force_control_switching_matching_2024, dipcom, dexforce, noseworthy_forge_2024, kobayashi_alpha-biact_2024, wu_tacdiffusion_2024, jaf}, with minor gravity compensation \citep{liu_forcemimic_force_control_2024}, with a fast Fourier transform to encode high-frequency force data as a 2D spectrogram \citep{acp}, or with an MLP \citep{factr, he_foar_2024, zhou_admittance_2024} into the observation space, and still yield effective tactile robot policies that appropriately act on force inputs. 

When the objective is to map sensor data to physical sensations, force data fundamentally provides a more direct and interpretable signal than visuotactile sensing, presenting promise for long-horizon and physically intricate tasks. While current research primarily applies force data to relatively short tasks such as grasping or pouring, its compact representation may help reasoning about prolonged and complex contact interactions, where visuotactile data may prove unwieldy or obfuscating.

Some works already condition on force to either select or modulate modes of action. He et al. post-process ground-truth future contact state from RGB images and whether contact force exceeds a manually-selected threshold value to train a contact predictor \citep{he_foar_2024}. This predictor then supervises a weighted prediction loss summed with the behavior cloning loss, which enables the policy to attend to force appropriately during and outside of contact. They additionally implement a reactive positon controller that conditions again on force to set more aggressive goal positions if insufficient F/T is detected during contact. Both the contact-supervised loss and force-conditioned reactive control enable better coverage and consistency in tasks like dry-erase board wiping, cucumber peeling, and pepper chopping. 

Liu et al. \citep{liu_forcemimic_force_control_2024} similarly learn to switch between free-space position control and force control conditioned on force feedback, and Noseworthy et al. \citep{noseworthy_forge_2024} learn to terminate a skill at either a specified or learned threshold. Outside of force data, Liu et al. (audio) \citep{maniwav}, Mejia et al. (audio) \citep{mejia_hearing_2024}, Feng et al. (optical and audio) \citep{msbot}, and  Li et al. (optical and audio) \citep{seehear} utilize multisensory self-attention to learn cross-modality, cross-time, and cross-modality-time relationships. This self-attention mechanism enables learning of adaptive weights for features at different task stages for action generation, resulting in greater task success and interpretability.

\subsection{Scaling Multimodal Reasoning}
While the trained FuSe policy in Jones et al. \citep{fuse} is largely limited to grasping, as discussed in \ref{sec:large_data}, it is also the only tactile robot policy thus far which finetunes a pretrained ``robot foundation model" backbone (Octo \citep{octo}). It is the pairing of the large, multimodal collected data and this pre-trained Octo policy which enable complex, generalizable reasoning about tactile properties, which we expand upon here.

As Octo is pretrained on a comparatively much larger dataset (OXE data \citep{openx}), Jones et al. identify that fine-tuning such a large pretrained model on novel tactile modalities with a “naive” mean-squared error (MSE) behavior cloning loss results in over-reliance on pre-training modalities such as camera vision and robot position data. Thus, Jones et al. design two multimodal losses which address this issue, using language, e.g. a task instruction such as ``pick up the squishy object'', as the ``glue'' across modalities. Each collected trajectory can have multiple task instructions: picking up a button can potentially alternate as one of picking up a (hard, metallic, red, circular) object. 

The first loss term is a contrastive loss to maximize mutual information between different modalities and semantics of the same scene. First, they construct an observation embedding from passing all modalities through the pre-trained Octo transformer and a multi-head attention layer. They compute a contrastive loss between each possible task instruction with this embedding and obtain an average $L_{contrast}$. 

The second loss term is a generative loss to learn high-level semantics for each possible combination of modalities, for which they build an embedding via the same process as above. Then, the embedding is passed through a generative head and a generative loss $L_{gen}$ is computed between the head output and ground truth language. During training, these auxiliary terms are summed to the MSE loss. Jones et al. show these multimodal losses enable compositional reasoning about tasks such as ``pick the object that has the same color as the button that plays piano."

The trained FuSe policy leverages tactile sensing and multimodal reasoning for discrete tasks like the given example, centered around classifying perceived objects or selecting objects whose predicted tactile properties correspond to a task instruction. Future works which incorporate force sensing, in addition to the visuotactile and audio finger sensing leveraged here, may be able to train tactile robot policies capable of both discrete and continuous tasks. The original Octo policy was also adapted for precise peg-insertion via fine-tuning on a small dataset of 100 demonstrations with wrist F/T sensing, showing downstream adaptability of their foundation model \citep{octo}, but no work has fully explored first-class large force pretraining for a tactile robot foundation model which can perform high-level (semantic reasoning) and low-level (reactive control) decision making for contact-rich, forceful tasks.

\section{Discussion: Towards Tactile Robot Foundation Models}\label{sec:wrapping_up}
In this article we reviewed 25 state-of-the-art works which train tactile robot policies mapping tactile sensing to robot actions in order to complete various contact-rich, forceful tasks. First, we explored the large space of data collection methods, highlighting methods which reduced the human-robot embodiment gap in sensing touch and issues related to scaling collection, human demonstrations, and system design
(Sec. \ref{sec:data_collection}). Then, we described the action space of tactile robot policies, drawing a line between position control and various force control methods, highlighting that low-level force control helped to bridge the human-robot embodiment gap and enabled fundamental physical skills (Sec. \ref{sec:action_space}). Finally, we discussed representation learning methods for force and visuotactile data, as well as methods for multimodal reasoning about touch in complex, multi-part tasks, highlighting opportunities to leverage force in large scale pretraining for discrete and continuous decision-making (Sec. \ref{sec:policy_learning}). These research areas each present highly relevant problems to overcome in order to build tactile robot foundation models and domestic robots which are suitably capable, safe, and versatile for human care.

\emph{On whether explicit force representations are needed:} The elephant in the room is whether explicit representations for force are actually needed or not. From a physiology perspective, there is evidence that neither tactile sensing or proprioception are strictly needed to implement dexterity and controlled motion. We argue that this kind of functional replacement simply demonstrates the large degree of redundancy that supports the human sensory-motor system, but should not be used to construe an argument that vision alone is sufficient for reliably functioning at high performance. 

From a controls perspective, impedance control does provide a pathway in which force can be actively controlled, yet does not need to be implicitly presented as an input to a foundation model. Impedance control is a low-level functionality of many robotic arms, and many contact-rich tasks might be solved by simply inferring appropriate mass, spring and damping parameters from task context and otherwise rely on position control. Here, impedance control is not limited to joint-torque sensors, but can include tactile sensing at the finger tips. Yet, impedance control might only be a subset of the various ways that end-effectors, including individual fingers, should react to external forces, and explicit representations of force might still be necessary for state representation. 

From a mechanical design perspective, sensing touch is not necessary for complex in-hand manipulation \citep{bhatt2022surprisingly}. Yet, sensing and impedance control is often implicit in soft mechanisms, and geometry and material choice are all critical in the open loop policies of \citep{bhatt2022surprisingly} to succeed. In \cite{yin2025dexteritygen}, basic impedance control is used to perform a variety of complex in-hand manipulation tasks, forgoing exact position control, a principle the authors refer to as implicit touch sensing. However, this method is not fully versatile, particularly for fine-grained manipulation tasks, and the authors propose future work to include actual touch sensing. 

\emph{On compositional vs. end-to-end policies:} Throughout this article, we have implicitly presumed that improving tactile robot policies learned end-to-end will eventually progress to generalist humanoid robots, while conceding and showing that for many of the reviewed works, the bounds of either ``end" varied. One could potentially argue that policies which produce force controller parameters, rather than exact robot motions, are not fully end-to-end and rather compositional methods. Commercially, Physical Intelligence's $\pi_0$ \citep{pi0}, Figure AI's Helix \citep{figure_helix}, and Google DeepMind's Gemini Robotics \citep{gemini_robotics} robot foundation models all leverage compositional control, in which a low-frequency model perceives the world and decomposes high-level task instructions into atomic skills for a high-frequency model to ingest and generate low-level control for. Other works explore even greater decomposition, leveraging large pretrained models to do step-by-step reasoning about object detection, picking appropriate grasp locations, judging the physical feasibility of actions, and determining appropriate modes of control \citep{ecot, metacontrol, reasoning2024arxiv_rad}. These compositional approaches, though complementary with tactile robot policies, suggest that robot foundation models can be functionally equivalent to several smaller and specialized policies.

By each focusing only on a few distinct tasks, tactile policies lag behind recent large robot models in generalizability across tasks, scenes, embodiments, and objects. This in part due to the nature of contact-rich manipulation which necessitate tactile sensing and often simply cannot be captured by the action vocabulary of such models. Additionally, the relative paucity and heterogeneity of tactile robot data currently precludes traditional large model pretraining techniques. Such issues are not mere engineering obstacles. They are representative of the fundamental phenomenon of physical sensing, which is combinatorial in input, processing, and output. 

The large variety of forces and task completion times spanning two orders of magnitude and resulting diversity in sensing modalities make it tempting to treat examples at the extreme ends of the spectrum as distinct problems. This is misleading, however, as reliable and efficient execution of these tasks might indeed require operation across the full spectrum. For example, dispensing accurate quantities from a jug of liquid will require precise force measurement across a large spectrum. Similarly, manipulation of delicate objects requires both high-level planning (in the orders of seconds) and high-frequency feedback control. Furthermore, in the interest of generalist platforms such as humanoids, foundation models will likely need to cover the entire spectrum of bandwidth in sensing, actuation, and control.

\emph{On scaling data collection:} There is no superlative method to collect forceful robot interaction data. Yet with contact-rich, tactile tasks being difficult to simulate with high fidelity, real-world data collection remains essential. Kinesthetic teaching, bilateral control, and handheld grippers improve upon teleoperation and each offer distinct trade-offs in terms of data fidelity, scalability, and human effort. Data collection of tactile robot data is an active research problem, and the research community remains in an exploratory phase where developing new, practical data collection methods is as important as refining existing methods.

While the current paradigm of data collection leans heavily on human demonstrations, history suggests that major breakthroughs in learning arise when we eliminate human bottlenecks, such as unsupervised learning of next-token prediction from large text corpora in natural language processing (NLP) \citep{brown2020gpt3}. As research progresses, the next leap may come not from refining human-driven data collection, but from unlocking scalable, automated methods that reduce dependency on human effort altogether. 

TacDiffusion learns precise peg insertion from pre-designed behavior tree controlled trajectories which do require expert design \citep{wu_tacdiffusion_2024}. Xie et al. \citep{jaf} learn delicate grasp policies also from a pre-designed adaptive grasp controller, but leverages LLM-supervision to parameterize the force controller, enabling automated versatility and generalizability across objects \citep{dg}, which follows a new proposed paradigm of distilling large (vision) language model guided robot trajectories into smaller robot policy models \citep{scalingup}.

Regardless of which method one chooses, the priority should be producing as much high-quality data as possible. This data leads to richer learning signals, better performing models, and ultimately, convergence toward effective tactile policy learning paradigms.

\section{Conclusion}
Imbuing transformer and diffusion-based end-to-end learning models with the ability to sense and generate forces is a recent trend, which builds up on a long history of force control in robotics. Adding force either as an input, an output, or both shows consistent improvements in makespan and robustness. While consistent with human physiology where tactile sensing and proprioception greatly improves dexterity, this also demonstrates that neither sense is critical and can be compensated by other modalities. The latter fact is motivation enough for pursuing force-less policies, particularly in development of minimalist, affordable robots which may be adequate for a subset of applications.

An intermediate solution between actively employing force sensing is the use of impedance control. Here, force control is only implicitly represented in a higher level controller, thereby reducing the data requirements during training. As impedance control is the computational equivalent of a mechanical spring and damper, it can also implemented as such, for example using soft, compliant actuators, also known as ``soft robotics'', or combinations of computational and mechanical compliance. Impressively demonstrated by biological systems, this mixed approach occupies a niche in robotics \cite{mengucc2017will} and is little explored in the context of robot learning. 

Only few of the works emphasize the saliency of tactile information, which can provide low-dimensional and even binary information on critical events such as contact. With even contact-rich tasks like in-hand manipulation being achieved via open-loop control or double-digit absolute improvements on benchmark tasks such as cloth-folding being achieved by model architecture improvements, we believe that the community has not really begun exploring highly dynamic tasks that cannot be solved without tactile sensing. We posit that tackling these tasks will require improving our understanding on representing forces and lead to models that are more data-efficient, cognizant of the physical world, and thus scalable and suitable for widespread adoption.

\bibliography{prelim}

\end{document}